\documentclass{article}
\usepackage{arxiv}
\usepackage[utf8]{inputenc} 
\usepackage[T1]{fontenc}    
\usepackage{hyperref}       
\usepackage{url}            
\usepackage{booktabs}      
\usepackage{amsfonts}       
\usepackage{nicefrac}       
\usepackage{microtype}     
\usepackage[norelsize]{algorithm2e}
\usepackage{multirow}
\usepackage{tabulary}
\usepackage{graphicx}
\usepackage{subcaption}
\usepackage{textcomp}
\title{Self Training Autonomous Driving Agent}

\author{
  Shashank Kotyan, Danilo Vasconcellos Vargas and Venkanna U.
  \\
  \texttt{\{shashank15100, venkannau\}@iiitnr.edu.in and vargas@inf.kyushu-u.ac.jp} \\
}

\begin{document}
\maketitle

\begin{abstract}
Intrinsically, driving is a Markov Decision Process which suits well the reinforcement learning paradigm. In this paper, we propose a novel agent which learns to drive a vehicle without any human assistance. We use the concept of reinforcement learning and evolutionary strategies to train our agent in a 2D simulation environment. Our model\textquotesingle s architecture goes beyond the World Model\textquotesingle s by introducing difference images in the auto encoder. This novel involvement of difference images in the auto-encoder gives better representation of the latent space with respect to the motion of vehicle and helps an autonomous agent to learn more efficiently how to drive a vehicle. Results show that our method requires fewer (96\% less) total agents, (87.5\% less) agents per generations, (70\% less) generations and (90\% less) rollouts than the original architecture while achieving the same accuracy of the original.
\end{abstract}

\keywords{Artificial Intelligence, Computer Vision, Deep Learning, Evolutionary Strategies, Reinforcement Learning}

\section{Introduction}\label{chapter:introduction}

Today, as the world is ushering into the era of automating things, one of key product of the industries which is yet to be automated completely is a vehicle which is also intelligent. According to Pcmag, "An autonomous vehicle is a computer-controlled car that drives itself" \cite{PCMAG}. There are many industries which are leading the research in autonomous vehicles, some of the most prominent are Google and Tesla. Research nowadays is inclining towards becoming leader in the new age of driver-less car. However, the current driver-less car is yet far from being the intelligent autonomous car. Current autonomous vehicles uses LiDAR (Light Detection And Ranging) technology, that measures distance to a target by illuminating the target with pulsed laser light and measuring the reflected pulses with a sensor. This measuring and analysis helps the autonomous vehicles to keep them in track. But, the various limitations associated with the LiDAR are: a) High dependency on prevention of object collision rather than driving. b) Range of LiDAR sensors is very low for a moving vehicle. c) Fails to work in the environment where the objects are in far range of the car. d) Needs guiding mechanism to take a turn, for example a curved object placed at the edge of the turn. e) Deployment of LiDAR in a consumer vehicle is expensive. \cite{LiDAR-1},\cite{LiDAR-2}
The new generation of autonomous vehicles, therefore are based on image analysis which is more realistic. This approach using vision benefits from image as raw input, to drive a vehicle. It helps in driving a vehicle more realistically in the perspective that as humans we use our sense of vision to drive rather than sense of touch (LiDAR). In the literature, there are various researches on the angle of the camera, some uses third person view, some uses first person view while some opt for bird eye view of the track. In the past decade, the research in the autonomous driving was dominated by supervised learning approach where a human expert\textquotesingle s data was made to be learned by the agent. Supervised Learning approach is more suitable for tasks which are based on a generalised formula (like classification and regression). As due to intrinsic property of supervised learning, it creates a function map of input to the output. A major limitation of supervised learning approach in this context, is requirement of dataset for driving which may or may not exist for all the environments in which an autonomous vehicle will be deployed. Also, creation of such datasets can be expensive and unfeasible, in some cases \cite{Supervised-2}. In supervised learning approach, an agent learns to imitate a human expert (strategy followed by human expert) instead of actually learning the best possible strategy in driving a vehicle.
Intuitively, driving is a Markov\textquotesingle s Decision Process (MDP) problem where a sequence of states are processed which involves the concepts of control theory. Though, for few smaller datasets generalised mapping of input and output works but at the same time it is not scalable or generalised enough for multiple environments to be deployed. Reinforcement Learning is a strategy to deal with Markov’s Decision Process problems. We learn the optimal strategy by sampling actions and then observing which one leads to our desired outcome. In contrast to the supervised approach, we learn this optimal action not from a classical label but from a time-delayed label called a reward. This scalar value tells us whether the outcome of whatever we did was good or bad. Hence, the goal of reinforcement learning is to take actions in order to maximize reward \cite{reinforcement}. Supervised Learning has two main tasks called Regression and Classification whereas Reinforcement Learning has different tasks such as exploitation or exploration, Markov’s decision processes, Policy Learning, Deep Learning and value learning \cite{Supervised-1}.
Research in reinforcement learning as taken a boost since the inception of deep learning. The reinforcement learning has been proved to overcome the supervised learning in a variety of control theory problems like Go and Atari Games. We developed an agent based on reinforcement learning to tackle the problem of driving. The motivation to develop the agent for autonomous vehicle stems from the experience of the authors who worked closely with the visually impaired people and their struggles in the daily life routines. A self-driving car will indeed help the visually challenged people to commute to their destination hassle-free and safely.
The following paper is organised as follows, Section \ref{chapter:related} covers the related works that has been done in the area of self-driving cars and the reinforcement learning, Section \ref{chapter:proposed} explains about the proposed method and algorithms used to develop the self-training autonomous driving agent, Section \ref{chapter:experiments} covers the experiments conducted for testing the proposed method with the state of the art, Section \ref{chapter:result} analyses the results from the experiments conducted for the proposed method and discusses it. Section \ref{chapter:conclusion} concludes the paper by iterating over the findings in the project. 

\section{Related Works}\label{chapter:related}
This section covers the existing agents present in the literature for autonomous driving and their brief summary about the contribution to research in self-driving cars and autonomous vehicles.
In the literature, there are various types of simulation tools available for training an agent for autonomous driving. Some of them are Open AI\textquotesingle s Car Racing \cite{openai}, Udacity\textquotesingle s Self Driving Car Simulator \cite{udacity}, Carla \cite{carla}, Torcs\cite{torcs}, etc. While all the simulators have their own advantages and disadvantages, we found that testing on Open AI\textquotesingle s car racing environment to be much simpler than the others due to a 2D environment rather than a 3D environment filling with graphics.
In 2015, using Deep Deterministic Policy Gradient (DDPG) Algorithm, Google was able to replicate and enhance the results of autonomous driving using LiDAR technology simulation. In the experiment phase they tested the algorithm on Torcs environment and were able to achieve the state of art results for the environment. While the approach was to use the sensory data simulation, like distance from the center of track, speed of car in direction of the track and orthogonal to track, it failed to solve the environment with images as raw input as it has solved other gym environments of Mujoco. Their approach was to use the actor-critic method for continuous control which took the benefits of both actor-critic algorithms and Q-Learning.\cite{ddpg}
Koutnik, et. al in their research showed the initial learning for a driver agent using rewards for simple tracks. In their approach, they relied on edge detection mechanisms of convolution networks to determine the location of the car. They showed that reinforcement learning can be applied to difficult problems involving images as input. Their core strategy was to use evolutionary algorithms to optimise the episodic reward for the agent. \cite{koutnik1},\cite{koutnik2} 
In many machine learning and deep learning applications, from age classification to image generation, it is a common practice to discretise continuous space. One intuition behind it argues that classification loss such as cross-entropy loss can send out more clear training signal than mean square error loss, especially if the outcomes show signs of clustering. This strategy that is used to make driving agent learn by discretising the values of steering angle, acceleration and brake opens up the realm of traditional reinforcement learning algorithms like Q-Learning and actor critic methods. Though, there is loss of some values associated but with discretisation. These algorithms tend to perform better than their continuous domain counter parts like DDPG algorithm.
More recently, world models architecture has solved the previously unsolved CarRacing environment in the continuous domain. This architecture takes benefit of the temporal features using a MDN-RNN module and optimise the episodic rewards using evolutionary strategies. Our model uses a modified architecture of the world models to boost the performance of the original architecture.

\section{Proposed Model}\label{chapter:proposed}
This section covers the proposed agent developed for the driving activity and the underlying algorithms present in the model. 

\subsection{Difference Image}\label{section:difference}

We define our difference image as the background subtraction between the two consecutive frames. This difference image also contains an interesting property of capturing the motion of the foreground and hence we believe using difference image, the underlying neural network will focus more on the features of the foreground than the background.
An auto encoder is defined in the literature as a special type of deep neural network whose primary objective is to find mappings for reconstruction in the given input domain. This type of deep neural network contains two major modules namely, Encoder, and Decoder. While the primary aim of the encoder is to learn a deterministic non-linear mapping of the given input domain to a lower-dimensional feature vector representation which is termed as latent space, the primary task of the decoder is to learn the inverse mapping of the latent space to the given input domain. To summarise, an auto encoder is a type of neural network which converts the input into a lower-dimensional latent space which is then used to reconstruct the input. Over the years, various types of auto encoders are researched upon mainly Variational Auto Encoder (VAE) and Stacked De-noising Auto Encoder (SDAE), Generative Auto Encoder (GAE), Relational auto encoder (RAE), and others. While the applications and the purpose of different auto encoders differ but they all are based on the same fundamental principle of reconstruction and creation of latent space.
Traditionally, the auto encoders’ neural network consisted of vanilla deep neural networks. But since, the introduction of convolution neural network and their proven superiority over vanilla deep neural networks for handling the image or multi-dimensional data, the current state-of the art auto encoders consist of convolution layers. We also use the convolution layers to extract the features from the image. However, we employ a hybrid approach of using convolution layers and the fully connected layers rather than creating an all-convolution model. The reason being, to better map the features to the latent space.

\subsection{Covariance Matrix Adaptation Evolution Strategy (CMA-ES)}\label{section:CMAES}
CMA-ES is a optimisation algorithm which is based on a class of algorithms classified as evolutionary algorithms. The CMA-ES algorithm is used to train the controller of the agent to find the best possible strategy for the controller to drive the vehicle. CMA-ES is suited for difficult non-linear non-convex black-box optimisation problems in continuous domain of which driving as an activity also forms a part. \cite{cmaes} 

\subsection{Architecture of Proposed Agent}
\begin{figure}
  \centering
  \includegraphics[width=.99\linewidth]{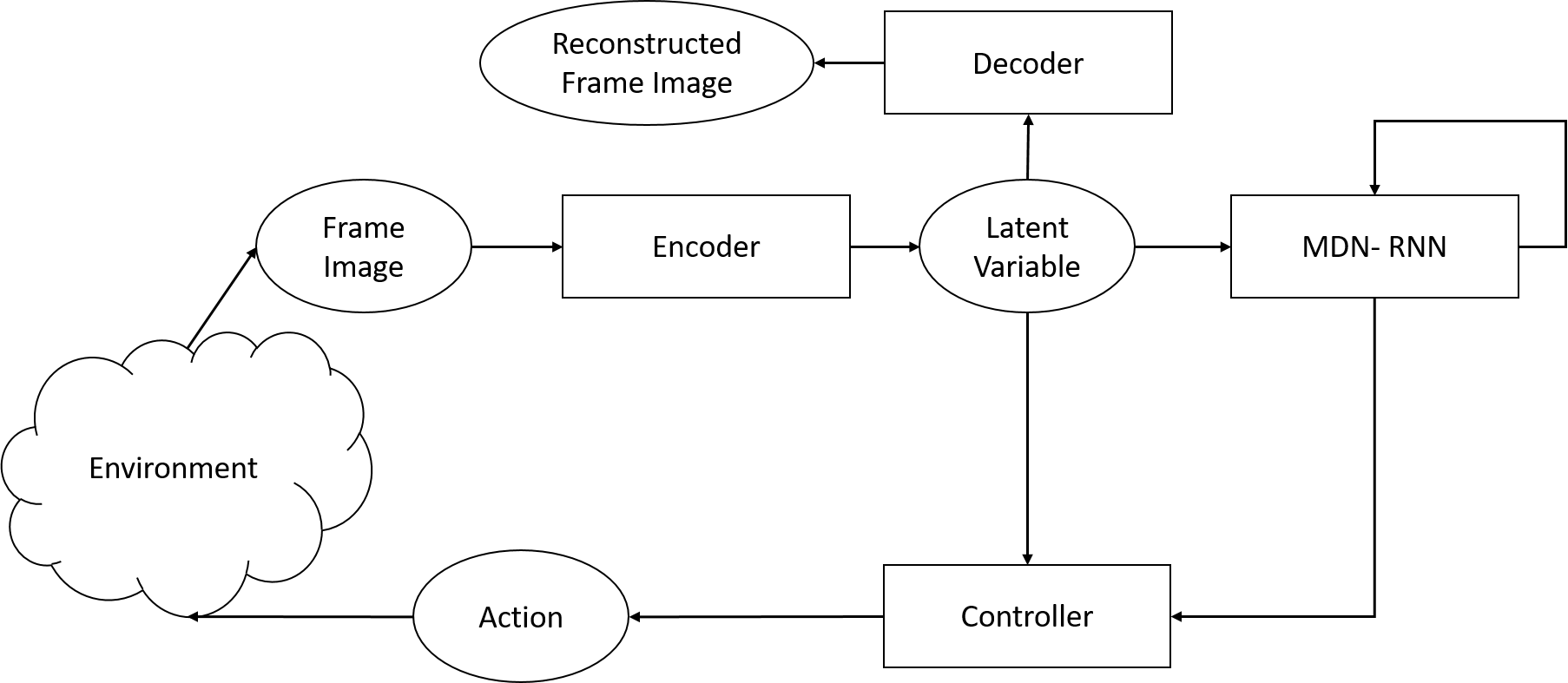}
  \caption{Architecture of STAD}
  \label{figure:worldmodels}
\end{figure}

Our proposed architecture have the three major components involved as shown in Figure \ref{figure:worldmodels}. The difference auto encoder analyses the image and reduces the dimension of the image for better processing in the further module of MDN-RNN. MDN-RNN captures the temporal properties of the frames associated with is required for the resolution of features which are temporally connected in the frames. The result of the MDN-RNN goes to the controller module which is responsible for the final action taken by the agent. This type of stacked architecture of the modules is inspired by the World Models Architecture presented by Ha et. al. \cite{worldmodels}. World Models Architecture is the current state-of-the art reinforcement learning architecture for the Open AI Car Racing environment. Our proposed agent introduces the difference image in the original architecture to give a boost in performance. 

\subsubsection{Difference Auto Encoder}
\begin{figure}
  \centering
  \includegraphics[width=.99\linewidth]{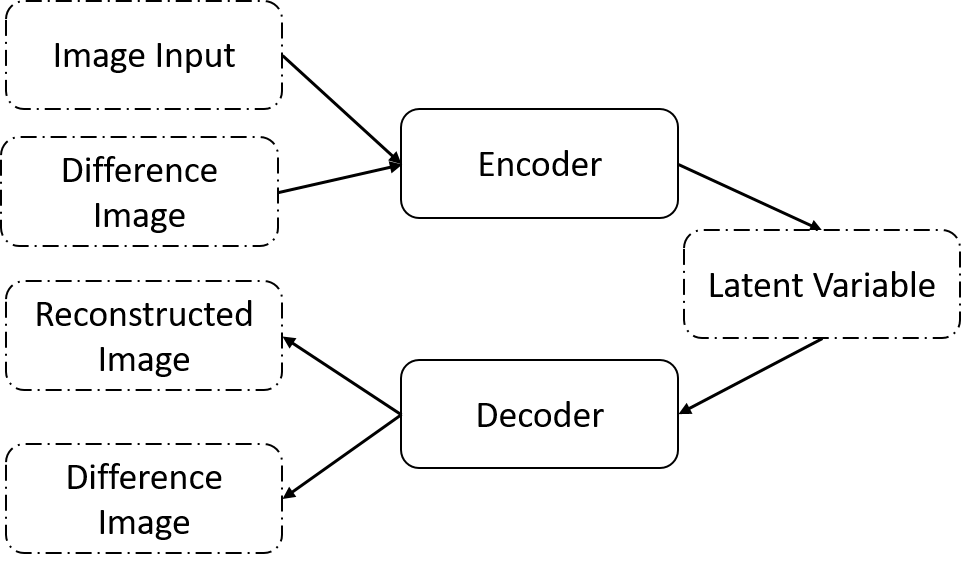}
  \caption{Auto encoder with original image and difference image as input and both original image and difference image as output}
  \label{figure:diffinputoutput}
\end{figure}

\begin{figure*}
  \centering
  \includegraphics[width=.99\linewidth]{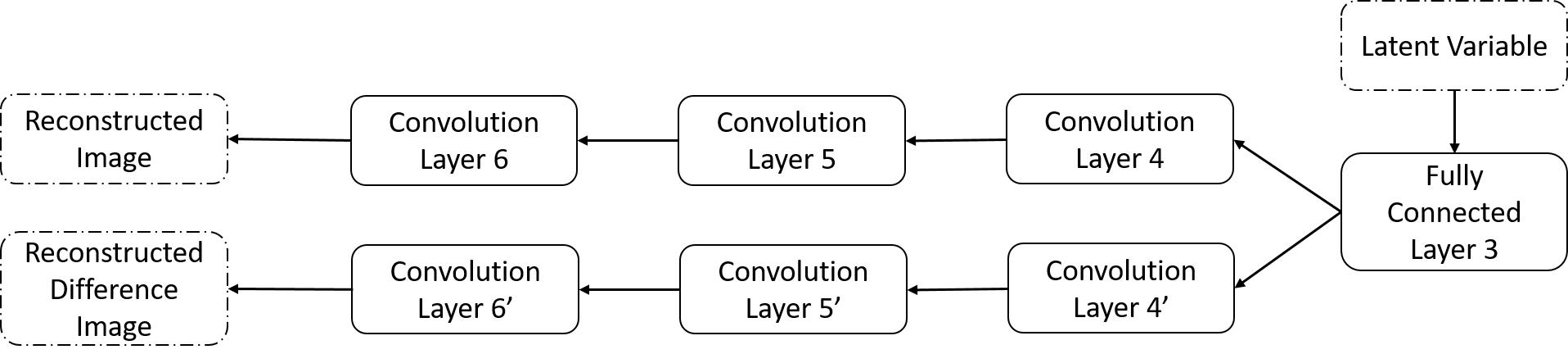}
  \caption{Architecture of Decoder with both original image and difference image as output}
  \label{figure:diffdecoder}
\end{figure*}

We define our difference auto encoder as a sub-class of auto encoder which involves the difference image processing and analysis merged into the latent space as shown in Figure \ref{figure:diffinputoutput}. To the best of our knowledge, this is the only architecture of auto encoder which deals with the construction of difference image from the latent space. The advantage of using difference image intuitively is to give neural network an explicit instruction to extract features from the foreground more aggressively than the background. This aggression towards the features of foreground not only helps in localising the region of interest but also helps in expressing the more vivid features in the latent space which gets neglected in the traditional auto encoders’ setup.  There can be three types of auto encoders possible with the involvement of difference image, 
\begin{itemize}
  \item Auto encoder with original image and difference image as input and only full image as output (Difference (Input) Auto Encoder)
  \item Auto encoder with original image and difference image as input and both original image and difference image as output (Difference (Input and Output) Auto Encoder)
  \item Auto encoder with only original image as input and both full image and difference image as output (Difference (Output) Auto Encoder) \\
\end{itemize}
Architecture of the the encoder and the decoder modules are shown in following figures. Figure \ref{figure:diffencoder} shows the architecture of the encoder involving both frame input and difference input and Figure \ref{figure:diffdecoder} shows the architecture of the encoder involving both frame and difference output.
 
\subsection{Auto Encoder}\label{section:autoencoder}
\begin{figure*}
  \centering
  \includegraphics[width=.99\linewidth]{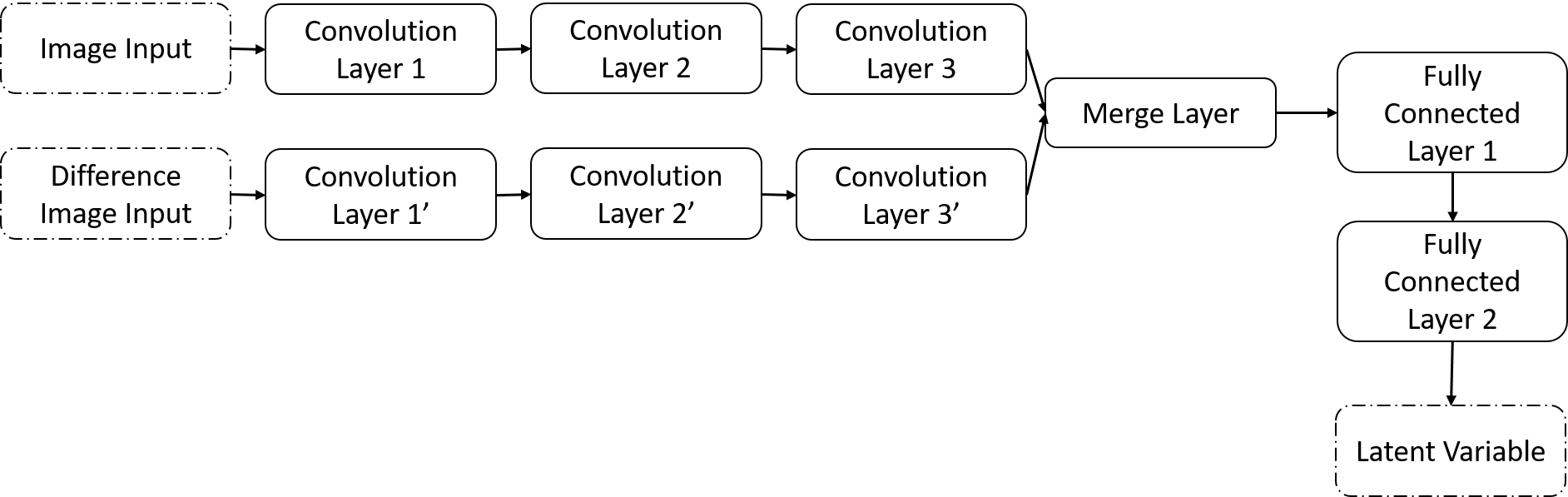}
  \caption{Architecture of Encoder with both original image and difference image as input}
  \label{figure:diffencoder}
\end{figure*}

\subsubsection{Recurrent Neural Network with Mixed Density Function (MDN-RNN)}

We define our recurrent neural network as a temporal feature extractor from the latent space. This module takes the input from the difference auto encoder in the form of latent space and extracts the temporal features which is fed to the controller.  

\subsubsection{Controller}

This module takes the input from the MDN-RNN to determine the steering angle, acceleration, and brake, for the agent to drive. This module is optimised and trained using the evolutionary algorithm CMA-ES to globally optimise the output of the agent.

\section{Experiments}\label{chapter:experiments}
The following section covers the details of the experiments conducted for the STAD agent.
The experiment wa conducted in the CarRacing-v0 environment developed by OpenAI. In the environment, the tracks are randomly generated for each trial, and the agent is rewarded for visiting as many tiles as possible in the least amount of time. The agent receives -0.1 for every frame and +1000/N for every track tile visited where N is the number of tiles in the track. Open AI describes the environment to be solved if the track is visited in 1000 frames i.e, with a reward of 900 (1000 - 0.1*1000). The environment is created in Box2D environment which gives an RGB image of the frame as the output and take a vector input of steering, acceleration and brake to drive the car. The camera position of the environment is the bird-eye where e get the top view of the environment focusing on the car. 
For training the variants of difference auto-encoder, a random roll-out of 1000 episodes were taken with 100 time frames. The difference image, if required was created with the help of every frame with four subsequent frames in the episode. This reduced the training images to 996 per episodes in case of difference images. 
For training the MDN-RNN module, an input vector containing the latent variable of the encoded frame and the random actions associated with it were fed to the module. For training the controller module, CMA-ES algorithm was used, a generation consisted of 8 agents was used. 
The system which was used to test the agent has the following configurations, 8 core processor, 16GB RAM and Nvidia GT710 Graphics Card. While the training of Difference Auto-encoder and MDN-RNN module was done with the help of Nvidia GT710, the training of controller was done using only the cores of the processor.
For testing, the model made to run for 100 random roll-outs and the average episodic reward in the 100 episodes was recorded. 

\section{Result and Analysis}\label{chapter:result}
\begin{figure}
  \centering
  \includegraphics[width=.99\linewidth]{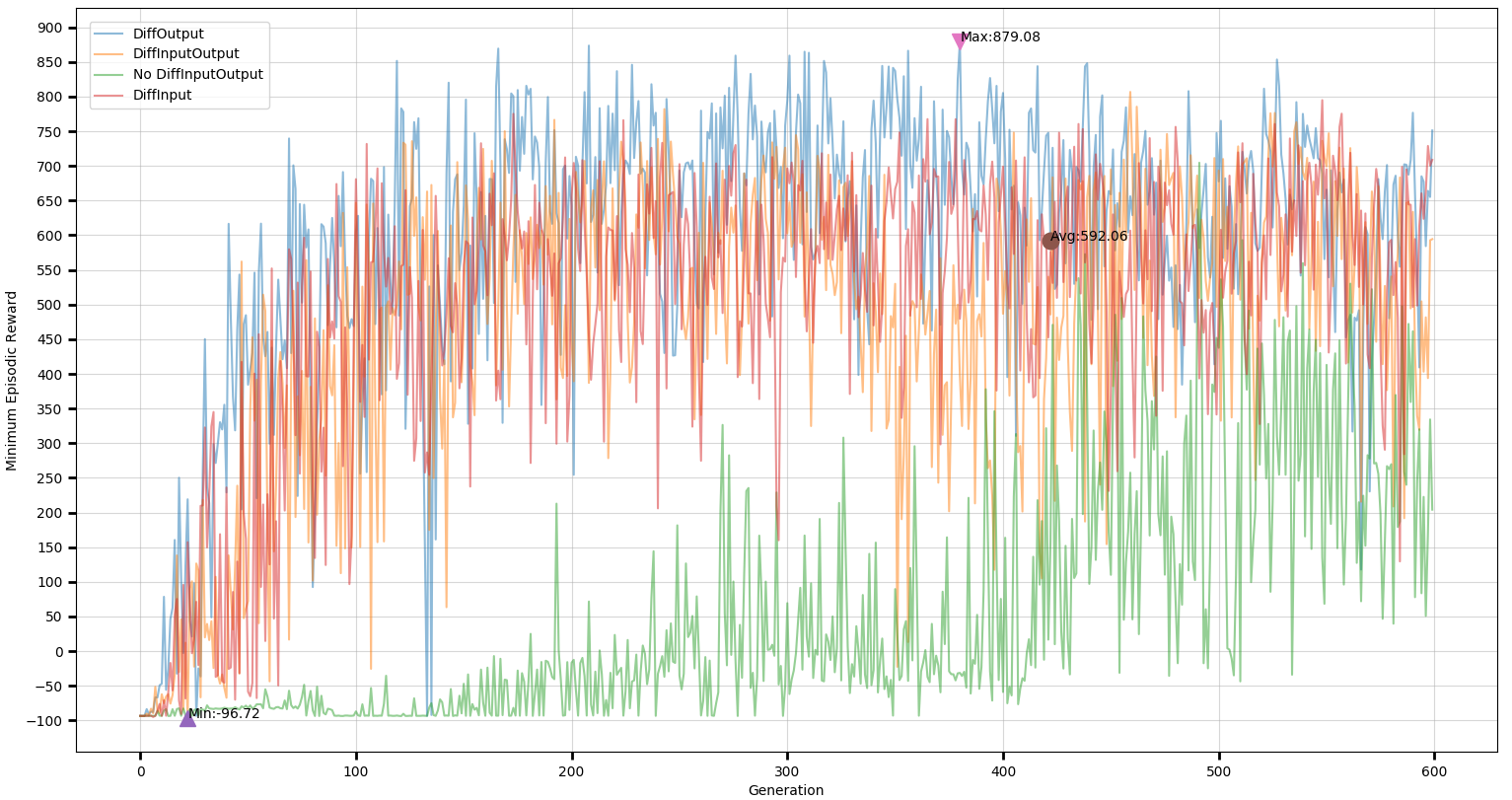}
  \caption{Performance Analysis of Worst Performer in the generation for CarRacing-v0 environment.}
  \label{figure:minimum}
\end{figure}

\begin{figure}
  \centering
  \includegraphics[width=.99\linewidth]{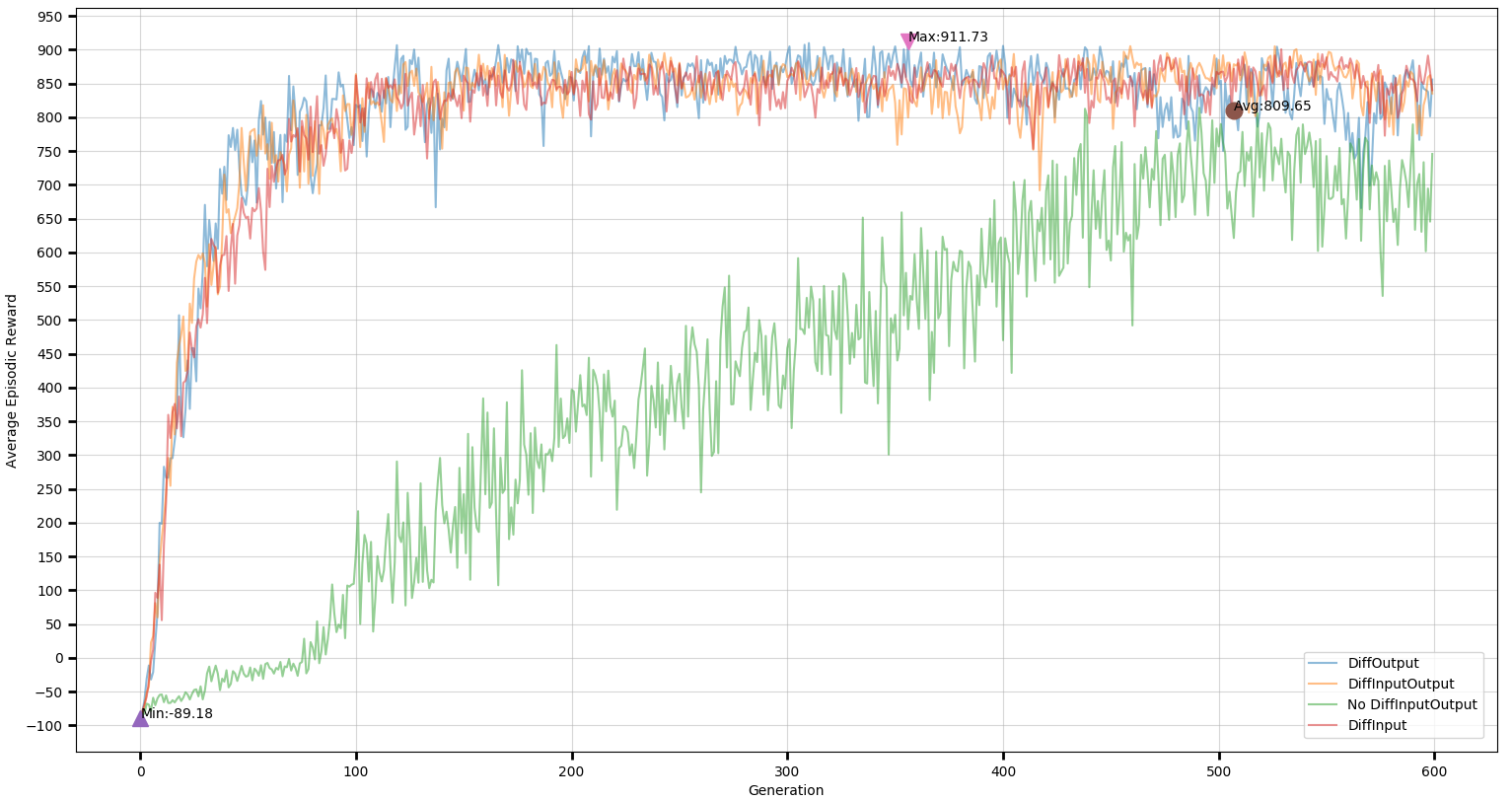}
  \caption{Performance Analysis of Average Performer in the generation for CarRacing-v0 environment. }
  \label{figure:average}
\end{figure}

\begin{figure}
  \centering
  \includegraphics[width=.99\linewidth]{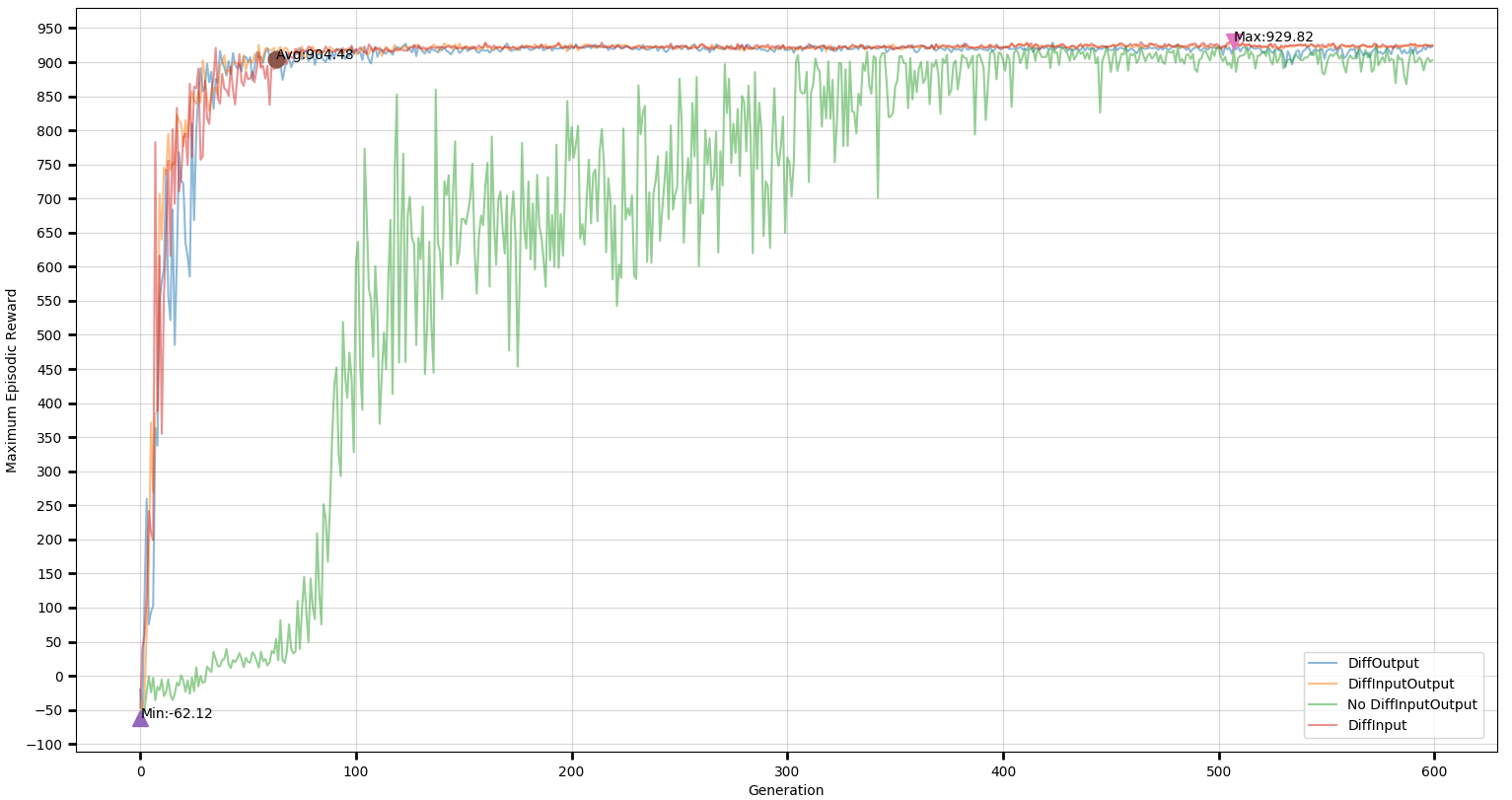}
  \caption{Performance Analysis of Best Performer in the generation for CarRacing-v0 environment. }
  \label{figure:maximum}
\end{figure}

Our agent with difference images is able to achieve the average score of 900+ episodic reward, effectively solving the task and obtaining similar results to the state of the art. Our agent achieves this average reward in a span of training for only 600 generations with only 1000 roll-outs compared to 2000 generations with 10000 rollouts in the original paper of world models. We have also used 8 agents per generations compared to 64 agents in the original paper. The result from the experiments shows the involvement of difference images boosts the performance of the architecture by effectively solving the environment in considerably less training. Numerically, our controller was trained on less (96\% )  total agents, (87.5\% ) agents per generations, (70\%) less generations and (90\%) rollouts used than the original paper to achieve similar results.  Table \ref{table:rewards} shows the performance comparison with other reinforcement learning algorithms and Table \ref{table:params} shows the difference in training parameters with World Models.
Figure \ref{figure:minimum} represents the minimum or worst reward a member of 8 population in the generation receives. Figure \ref{figure:average} represents the average reward a member of 8 population in the generation receives. Figure \ref{figure:maximum} represents the maximum or best reward a member of 8 population in the generation receives. With the episodic rewards results, we can say that the adding difference image to the model boosts the performance of the architecture. X-axis of the figures represents generation and y-axis represents episodic reward for the performer. Best performer in the generation, is the agent which has highest episodic reward in the generation while worst performer in the generation, is the agent which has lowest episodic reward in the generation. Average performer in the generation is average episodic reward of the generation\textquotesingle s agents. We were able to find best performer in the training with 929.82 episodic reward i.e, the track tile is visited in 702 frames. We find that with difference images, the best performer achieves stability relatively early then without difference images.We were also able to find average performer in the training with 911.73 episodic reward. However, in the 100 random rollouts of testing, the agents weren\textquotesingle t able to achieve those high results because of few poor episodic perfomances out of 100. The average episodic reward for top 90 test episodes was approximately 915-925.

\begin{table}
  \centering
  \begin{tabular}{l c}
    \hline
    Algorithm                       & Average Episodic Reward   \\
    \hline
    Task Aware Autoencoder \cite{taskautoencoder}   & 200             \\    
    DQN \cite{dqn}                  & 343             \\
    DDPG \cite{a3cc}                & 100             \\
    A3C (Continuous) \cite{a3cc}              & 591             \\
    A3C (Discrete)  \cite{a3cd}             & 652               \\
    Open AI Benchmark \cite{openaibench}      & 838               \\
    \hline
    World Model \cite{worldmodels}            & \textbf{906}          \\
    \hline
    World Model (Fewer Training)            & 827               \\
    STAD (Difference Input - Fewer Training )   & \textbf{905}          \\
    STAD (Difference Output - Fewer Training)     & \textbf{906}          \\
    STAD (Difference Input and Output - Fewer Training) & \textbf{902}          \\
    \hline
  \end{tabular}
  \caption{Episodic Reward for other Reinforcement Learning Algorithms in CarRacing-v0 Environment}
  \label{table:rewards}
\end{table} 

\begin{table}
  \centering
  \begin{tabular}{l c c}
    \hline
    Parameters              & Original\cite{worldmodels}  & STAD              \\
    \hline
                            & \multicolumn{2}{c}{For VAE and MDN-RNN}   \\
    Rollouts                & 10000     & 1000 (10\%)         \\    
    \hline
                            & \multicolumn{2}{c}{For Controller}    \\
    Number of Generations   & 2000    & 600 (30\%)          \\
    Agents per Generation   & 64    & 8 (12.5\%)          \\
    Total Agents Evaluated  & 128000  & 4800 (3.75\%)         \\
    \hline
  \end{tabular}
  \caption{Parameters used for training world models. 
  We have used only 10\% of the original rollouts, 30\% of the original generations, 12.5\% of the original agents per generation and 3.75\% of the original total agents.
  }
  \label{table:params}
\end{table} 

\section{Conclusion}\label{chapter:conclusion}
We proposed a novel agent for autonomous driving based on difference images. The agent\textquotesingle s architecture was inspired by World Models Architecture which is the current state-of-the art for car racing environment. Our results shows that the including difference image in the output and input, it boosts the performance of the state of the art and achieve the results of the state of the art in 96\% less training of the model. We show the importance of using difference image in the activities in which motion is involved. Our model also achieves stability in the reward by agents in population earlier than the state-of the art. Our models justifies the self-training as no human expert was involved in the training and assisting the agent. The agent learns to drive the car smoothly with the help of environment specified rewards.  \\
However, with the current involvement of difference images there is still a limitation of producing the same effect in the first person view of self-driving cars. In the real-life environment, an agent will rarely see the top-view of the car it handles. It will mostly be a third-person view or the first-person view. Therefore, it is necessary to develop an architecture which not only can handle the motion parameters in the still frame but also is invariant to the view of the agent. Our future work, will cover into the developing a model architecture which is view-invariant for the self-training cars. 

\section*{Acknowledgement}
This work was supported by JST, ACT-I Grant Number JP-50166, Japan.


\begin{thebibliography}{9}
  \bibitem{PCMAG}
  "self-driving car Definition from PC Magazine Encyclopedia". www.pcmag.com.
  \bibitem{LiDAR-1}
  Wei, J., Snider, J. M., Kim, J., Dolan, J. M., Rajkumar, R., \& Litkouhi, B. (2013, June). Towards a viable autonomous driving research platform. In Intelligent Vehicles Symposium (IV), 2013 IEEE (pp. 763-770). IEEE.
  \bibitem{LiDAR-2}
  Rasshofer, R. H., \& Gresser, K. (2005). Automotive radar and lidar systems for next generation driver assistance functions. Advances in Radio Science, 3(B. 4), 205-209.
  \bibitem{Supervised-2}
  Ye, C., Yung, N. H., \& Wang, D. (2003). A fuzzy controller with supervised learning assisted reinforcement learning algorithm for obstacle avoidance. IEEE Transactions on Systems, Man, and Cybernetics, Part B: Cybernetics.
  \bibitem{reinforcement}
  Sutton, R. S., \& Barto, A. G. (2018). Reinforcement learning: An introduction. MIT press.
  \bibitem{Supervised-1}
  Kaelbling, L. P., Littman, M. L., \& Moore, A. W. (1996). Reinforcement learning: A survey. Journal of artificial intelligence research, 4, 237-285.
  \bibitem{ddpg}
  Lillicrap, T. P., Hunt, J. J., Pritzel, A., Heess, N., Erez, T., Tassa, Y., ... \& Wierstra, D. (2015). Continuous control with deep reinforcement learning. arXiv preprint arXiv:1509.02971.
  \bibitem{koutnik1}
  Koutnik, J., Cuccu, G., Schmidhuber, J., \& Gomez, F. (2013, July). Evolving large-scale neural networks for vision-based reinforcement learning. In Proceedings of the 15th annual conference on Genetic and evolutionary computation (pp. 1061-1068). ACM.
  \bibitem{koutnik2}
  Koutnik, J., Schmidhuber, J., \& Gomez, F. (2014, July). Evolving deep unsupervised convolutional networks for vision-based reinforcement learning. In Proceedings of the 2014 Annual Conference on Genetic and Evolutionary Computation(pp. 541-548). ACM.
  \bibitem{cmaes}
  Hansen, N. (2016).The CMA evolution strategy: A tutorial. arXiv preprint arXiv:1604.00772.
  \bibitem{openai}
  Brockman, G., Cheung, V., Pettersson, L., Schneider, J., Schulman, J., Tang, J., \& Zaremba, W. (2016).Openai gym. arXiv preprint arXiv:1606.01540.
  \bibitem{udacity}
  Udacity. An open source self-driving car, 2017
  \bibitem{carla}
  Dosovitskiy, A., Ros, G., Codevilla, F., Lopez, A., \& Koltun, V. (2017).CARLA: An open urban driving simulator. arXiv preprint arXiv:1711.03938.
  \bibitem{torcs}
  Wymann, B., Espié, E., Guionneau, C., Dimitrakakis, C., Coulom, R., \& Sumner, A. (2000).Torcs, the open racing car simulator. Software available at http://torcs. sourceforge. net, 4, 6.
  \bibitem{worldmodels}
  Ha, D., \& Schmidhuber, J. (2018).Recurrent world models facilitate policy evolution. arXiv preprint arXiv:1809.01999.
  \bibitem{taskautoencoder}
  Liang, E., Fox, R., Gonzalez, J. E., \& Stoica, I. (2018). Task-Relevant Embeddings for Robust Perception in Reinforcement Learning.
  \bibitem{dqn}
  L. Prieur.Deep-Q Learning for racecar reinforcement learning problem. https://goo.gl/VpDqSw, 2017.
  \bibitem{a3cc}
  S. Jang, J. Min, and C. Lee. Reinforcement car racing with A3C. https://goo.gl/58SKBp, 2017.
  \bibitem{a3cd}
  M. Khan and O. Elibol. Car racing using reinforcement learning. https://goo.gl/neSBSx, 2016
  \bibitem{openaibench}
  O. Klimov.CarRacing-v0. http://gym.openai.com/, 2016.

\end{thebibliography}
\end{document}